\documentclass[10pt,a4paper]{article}
\usepackage{amssymb,amsfonts,amsmath}
\usepackage{amsthm,eucal}
\usepackage[utf8]{inputenc}

\usepackage{cite}

\usepackage{nameref,hyperref}

\usepackage{lastpage,fancyhdr,graphicx}
\usepackage{color}

\definecolor{Gray}{gray}{.25}
\newtheorem{theorem}{Theorem}
\newtheorem{lemma}[theorem]{Lemma}

\newcommand{\proba}{\mathbb{P}}
\newcommand{\R}{\mathbb{R}}
\newcommand{\E}{\mathbb{E}}
\newcommand{\Lcal}{\mathcal{L}}
\newcommand{\Fcal}{\mathcal{F}}
\begin{document}

\title{The convergence of the Stochastic Gradient Descent (SGD) : a self-contained proof}
\author{Gabriel TURINICI
\\
CEREMADE - CNRS \\ Universit\'e Paris Dauphine - PSL Research University 
\\
\texttt{Gabriel.Turinici@dauphine.fr}}
\date{\today}

% document begins here
\maketitle
\begin{abstract}
    We give  here a proof of  the convergence of the Stochastic Gradient Descent (SGD) in a self-contained manner.
\end{abstract}

\section{Introduction}
 The Stochastic Gradient Descent (SGD) or other algorithms derived from it are used extensively in Deep Learning, a branch of Machine Learning; but the proof of convergence is not always easy to find. The goal of this paper is to adapt various proofs from the literature in a simple format.  
 {\bf In particular no claim of originality is made and this is rather a pedagogic work} (see \cite{turinici2020convergence,ayadi2020stochastic,turinici2020xray,Turinici_2019
} for some of my recent research papers in this area); please cite this presentation if you find it useful.

This proof can be used in any domain where a self-contained presentation is needed.
 
 \section{Recall of the general framework}
 
 Suppose $(\Omega,F,\proba)$ is a probability space,  
 $L:\Omega \times \R^N \to \R$ a function depending on a random argument $\omega$ and a parameter $X$ (second argument) to be optimized. Denote 
 \begin{equation}
     \Lcal(X) = \E_\omega [ L(\omega,X)].
 \end{equation}
 The goal of the SGD is to find a minimum of $\Lcal$. It operates iteratively by taking at iteration $n$:
\begin{itemize}
\item  a (deterministic) "learning rate" $\rho_n$ (schedule fixed {\it a priori})
\item a random $\omega_n \in \Omega$ independent of any other previous random variables is drawn (following the law $\proba$)
\item and updating by the formula
\begin{equation}
X_{n+1} = X_n - \rho_n \nabla_X L(\omega_n,X_n).
\end{equation}
\end{itemize}

\section{Hypothesis on $L$ and $\Lcal$}
 
 In order to prove the convergence we need some hypothesis that are detailed below
 
\begin{enumerate}
\item 
    The gradient of $L$ satisfies the following bound:
 \begin{equation}     \exists C_0, C_1 > 0 : \      \E_\omega \left[ 
     \| \nabla_X L(\omega,X) \|^2
     \right] \le C_0 + C_1 \|X\|^2, \ \forall X \in \R^N.
 \label{eq:hypgradbounded}
 \end{equation}
% \begin{equation}
%     \exists B > 0 : \ 
%     \sup_\omega 
%     \| \nabla_X L(\omega,X) \|^2 \le B, \ \forall X \in \R^N.
% \label{eq:hypgradbounded}
% \end{equation}
\item $\Lcal$ is strongly convex:
\begin{equation}
\exists \mu > 0 : \     \Lcal(Y) \ge \Lcal(X) + \langle \nabla \Lcal(X), Y-X \rangle + \frac{\mu}{2} \|X-Y\|^2, \ \forall X,Y \in \R^N.
\label{eq:hypstrongconvex}
 \end{equation}
Note that for $\mu=0$ this is just the usual convexity, i.e. the function is above its tangent. 
% WWW TODO: plot tangent, parabola
For general $\mu$ this tells that the function is even above a parabola centered in any $X$. For regular functions this means that the Hessian
$D^2 \Lcal$ of $\Lcal$ satisfies $D^2 \Lcal \ge \mu \cdot I_N$\footnote{Here $I_N$ is the $N\times N$ identity matrix.}. 
\end{enumerate}

\section{A convergence result and its proof}

We fill prove the following

\begin{theorem}
Suppose that each $L(\omega,\cdot)$ is differentiable (a.e. $\omega \in \Omega$)\footnote{This requirement can be largely weakened. For instance in the case of ReLU activation, which corresponds to the positive part $x\mapsto x_+$,  one can employ any suitable sub-gradient of the $x_+$ function and in particular take at the non-regular point $x=0$ any value between $0$ and $1$.} and that 
$\Lcal$ satisfies the hypothesis 
\eqref{eq:hypgradbounded} and \eqref{eq:hypstrongconvex}. 
 Then 
 \begin{enumerate}
\item the function $\Lcal$ has an unique minimum $X_*$;
\label{item:uniqueness}
     \item \label{item:inequalitydn}
For any $n \ge 0$ denote
 \begin{equation}
     d_n =  \E \left[ \|X_{n}-X_*\|^2 \right].
 \end{equation}
 Then there exist constants $c_0, c_1 >0$ such that
 \begin{equation}
     d_{n+1} \le      (1-\rho_n \mu + \rho_n^2 c_1) d_n + \rho_n^2 c_0.
\label{eq:dnplusone} \end{equation}
\item \label{item:constant}
For any $\epsilon > 0$ there exists a $\rho_\epsilon > 0$ such that if $\rho_n = \rho < \rho_\epsilon$ then 
\begin{equation}
    \limsup_{n\to \infty}  \E \left[ \|X_{n+1}-X_*\|^2 \right] \le \epsilon.
\label{eq:estimationvoisinage}
\end{equation}
\item \label{item:convergence}
Take $\rho_n$ a sequence such that:
\begin{equation}
\rho_n \to 0 \text{ and } \sum_{n\ge 1} \rho_n = \infty.
\label{eq:hyprhon}
\end{equation}
 Then $d_n \to 0$, that is 
 $\lim_{n \to \infty }X_n = X_*$, where the convergence is the $L^2$ convergence of random variables.
\end{enumerate}
 \begin{proof}
{\bf Item \ref{item:uniqueness}}:
The existence and uniqueness of the optimum is guaranteed by the assumptions of strong convexity and smoothness of $\Lcal$.

\noindent {\bf Item \ref{item:inequalitydn}}:
We have
\begin{align}
&    \E \left[ \|X_{n+1}-X_*\|^2 \right]  = 
    \E \left[ \|X_{n}-X_* - \rho_n \nabla_x L(\omega_n,X_n)\|^2 \right]  
\nonumber    \\&= 
    \E \left[ \|X_{n}-X_*\|^2 \right]  +
\rho_n^2    \E \left[ \|\nabla_x L(\omega_n,X_n)\|^2 \right]  
- 2 \rho_n 
    \E \left[ \langle X_{n}-X_*, \nabla_X L(\omega_n,X_n)\rangle \right].
\end{align}
First we remark that\footnote{The formal justification is as follows: 
denote  by $\Fcal_n$ the sigma algebra generated by $X_1$, ..., $X_n$, $\omega_1$, ..., $\omega_{n-1}$. In particular $\omega_n$ is independent of
$\Fcal_n$. Recall now that for any random variables $U$ measurable with respect to $\Fcal_n$ and $V$ independent of $\Fcal_n$:
$ \E[ g(U,V) | \Fcal_n] = \int g(v,U) P_V(dv) $ and in particular
$\E[ g(U,V)]= \E[\E[ g(U,V) | \Fcal_n]]= 
\E[ \int g(v,U) P_V(dv)]$.
}
$$   \E \left[ \langle X_{n}-X_*, \nabla_x L(\omega_n,X_n)\rangle \right]
 = \E \left[ \langle X_{n}-X_*, \nabla \Lcal(X_n)\rangle \right].
$$
But at its turn
\begin{align}
    &
\E \left[ \langle X_{n}-X_*, \nabla \Lcal(X_n)\rangle \right] \ge 
\E \left[ \Lcal(X_n) - \Lcal(X_*) + \frac{\mu}{2}  \|X_{n}-X_*\|^2 \right] 
\nonumber\\& 
\ge \frac{\mu}{2}  \E  [\|X_{n}-X_*\|^2],
\end{align}
the last inequality being guaranteed by the fact that $X_*$ is the minimum.
Putting together all relations proved so far one obtains the relation
\eqref{eq:dnplusone} (we have used hypothesis \eqref{eq:hypgradbounded} to bound the term $\E \|\nabla_x L(\omega_n,X_n)\|^2 $ by $c_0+ d_n c_1$).

\noindent {\bf Item \ref{item:constant}}:
When $\rho_n$ is constant equal to $\rho$ inequality \eqref{eq:dnplusone} is equivalent to 
$$d_{n+1} -  \frac{\rho c_0}{\mu- \rho c_1} \le (1-\rho \mu+ \rho^2 c_1) \left(d_n - \frac{\rho c_0}{\mu- \rho c_1}\right).$$
Since the function $x \mapsto x_+$ (the positive part) is increasing we obtain for $\rho < \min(1/\mu, \mu/2 c_1)$:
$$\left(d_{n+1} - \frac{\rho c_0}{\mu- \rho c_1} \right)_+
\le \left(1- \frac{\rho \mu}{2}\right) \left(d_n - \frac{\rho c_0}{\mu- \rho c_1}\right)_+,$$
and by iteration, for any $k\ge 1$:
$$\left(d_{n+k} - \frac{\rho c_0}{\mu- \rho c_1} \right)_+
\le \left(1-\frac{\rho \mu}{2}\right)^k \left(d_n - \frac{\rho c_0}{\mu- \rho c_1}\right)_+.$$
Taking $k\to \infty$ we obtain 
$\limsup_k \left(d_{k} - \frac{\rho c_0}{\mu- \rho c_1} \right)_+=0$ hence the conclusion \eqref{eq:estimationvoisinage} for $\rho$ smaller than 
$\rho_\epsilon:=\min\{1/\mu, \mu/2c_1, \epsilon \mu / (c_0-\epsilon c_1)\}$.

\noindent {\bf Item \ref{item:convergence}}:
For non-constant $\rho_n$ and arbitrary fixed $\epsilon$
we obtain from \eqref{eq:dnplusone}
$$d_{n+1} - \epsilon \le \left(1-\frac{\rho_n \mu}{2}\right) (d_n - \epsilon) + \rho_n (c_0\rho_n - \mu \epsilon/2 + (\rho_n c_1 - \mu/2)d_n).$$ 
When $n$ is large enough the last term in the right hand side is negative and 
thus 
$$d_{n+1} - \epsilon \le \left(1-\frac{\rho_n \mu}{2}\right) (d_n - \epsilon),$$ 
therefore 
$$\left(d_{n+k} - \epsilon\right)_+
\le \left(1-\frac{\rho_n \mu}{2}\right) \left(d_n - \epsilon\right)_+.$$
Iterating such inequalities we obtain
$$\left(d_{n+k} - \epsilon \right)_+
\le \prod_{\ell=n}^{n+k-1} \left(1-\frac{\rho_\ell \mu}{2}\right) \left(d_n - \epsilon \right)_+.$$
From the Lemma \ref{lemma:product} we obtain
$\lim_{k \to \infty} \left(d_{k} - \epsilon \right)_+=0$ and since this is true for any $\epsilon$ the conclusion follows.
\end{proof}

\end{theorem}
\begin{lemma}
Let $\xi >0$ and $\rho_n$ a sequence of positive real numbers such that $\rho_n \to 0$ and $\sum_{n\ge 1} \rho_n = \infty$. Then for any $n \ge 0$:
\begin{equation}
\lim_{k\to \infty}    \prod_{\ell= n}^{n+k} (1-\rho_\ell \xi) = 0.
\end{equation}
\label{lemma:product}
\end{lemma}
\begin{proof}
Since $\rho_n\to 0$,  $\rho_\ell \xi<1$ for $\ell$ large enough. To keep things simple we suppose this is true starting from $n$. 
Recall that for any $x \in ]0,1[$ we have $\log(1-x) \le -x$; 
then:
\begin{equation}
0 \le   \prod_{\ell= n}^{n+k} (1-\rho_\ell \xi) = e^{ \sum_{\ell= n}^{n+k} \log(1-\rho_\ell \xi)} \le e^{ \sum_{\ell= n}^{n+k} (-\rho_\ell \xi)} \overset{k\to \infty}{\longrightarrow} e^{- \infty} =0,
\end{equation}
which concludes the proof.
\end{proof}

\section{Concluding remarks}
We make here some remarks concerning the hypothesis and the use in Neural Networks.

First, consider the hypothesis $\sum_n \rho_n= \infty$; at first it may seem strange but this is not really so\footnote{One may show on a simple counter-example $L(\omega,X) = (X-\omega)^2/2$ with $\omega$ a standard normal that $\sum_n \rho_n <\infty$ will lead to a non-null limit variance; to do so, use a second order version of lemma~\ref{lemma:product} and the formula $Var(X_{n+1}) = (1-\rho_n)^2 Var(X_n) + \rho_n^2$ true in this case.} . Note that in particular it is true when $\rho_n$ is a constant. But in general, if we forget the stochastic part\footnote{This can be made precise when the stochastic part is added, see \cite{ayadi2020stochastic}.}, one can interpret the SGD as following some continuous time dynamics  of the type $X'(t) = - \nabla \Lcal(X)$; for the simple quadratic function $\Lcal(X)=  \alpha \|X\|^2 / 2$ the dynamics is 
$X'(t)  = -\alpha X(t)$ with solution $X(t) = e^{-\alpha t}X(0)$ needing an infinite 'time' $t$ to converge to the minimum $X_*=0_N$. Or here $\sum_n \rho_n$ is the discrete version of the time and thus it is not a surprise to need infinite time to obtain $X_*$ with infinite precision. On the other hand if a finite precision is needed one can just take a constant time step as indicated in the theorem\footnote{but in this case one may spend a too long time to wait for the convergence to this small neighborhood to arrive see \cite{ayadi2020stochastic} for some ways to accelerate the convergence.}.

Note that an important example that satisfies 
\eqref{eq:hyprhon} is $\rho_n = \frac{c_3}{c_4 + n}$, with $c_3, c_4 > 0$. In general giving a functional form for $\rho_n$ is termed 'choosing a decay rate', but it may not be clear what the best decay rate is in general.

Finally, concerning the hypothesis \eqref{eq:hypgradbounded} and \eqref{eq:hypstrongconvex} both can be considerably weakened, but at the cost of a longer proof.

\section*{Acknowledgements}
A special thanks to Stefania Anita for helpful discussions concerning this work and in particular for suggesting the present form of the hypothesis \eqref{eq:hypgradbounded}.

%\bibliography{library}
%\bibliographystyle{plain}

\end{document}